\documentclass[10pt,twocolumn,letterpaper]{article}

\usepackage{cvpr}
\usepackage{times}
\usepackage{epsfig}
\usepackage{graphicx}
\usepackage{amsmath}
\usepackage{amssymb}
\usepackage{pgfplots}
\usepackage{setspace}
\usepackage{algorithm}
\usepackage{algorithmic}
\pgfplotsset{compat=newest}

\usepackage{xcolor}
\usepackage{graphicx}
\usepackage{subcaption}

\usepackage{mathtools}
\usepackage{ mathrsfs }
\usepackage{calrsfs}
\DeclareMathAlphabet{\pazocal}{OMS}{zplm}{m}{n}
\newcommand{\Lb}{\pazocal{L}}

\let\svthefootnote\thefootnote

\newcommand\blankfootnote[1]{%
  \let\thefootnote\relax\footnotetext{#1}%
  \let\thefootnote\svthefootnote%
}
\let\svfootnote\footnote
\renewcommand\footnote[2][?]{%
  \if\relax#1\relax%
    \blankfootnote{#2}%
  \else%
    \if?#1\svfootnote{#2}\else\svfootnote[#1]{#2}\fi%
  \fi
}

\usepackage[breaklinks=true,bookmarks=false]{hyperref}

\cvprfinalcopy 

\ifcvprfinal\fi
\begin{document}

\title{Deep Temporal Linear Encoding  Networks}

\author{
    {  Ali Diba$^{1,\star}$,  Vivek Sharma$^{1,\star}$, and Luc Van Gool$^{1,2}$}\\
    {\normalsize {$^{1}$ESAT-PSI, KU Leuven,  $^{2}$CVL, ETH Z\"{u}rich}} \\ 
     \tt\small \{firstname.lastname\}@esat.kuleuven.be  
 }


\maketitle

\begin{abstract}
{
The CNN-encoding of features from entire videos for the representation of human actions has rarely been addressed. Instead, CNN work has focused on approaches to fuse spatial and temporal networks, but these were typically limited to processing shorter sequences. We present a new video representation, called temporal linear encoding (TLE) and embedded inside of CNNs as a new layer, which captures the appearance and motion throughout entire videos. It encodes this aggregated information into a robust video feature representation, via end-to-end learning. Advantages of TLEs are: (a) they encode the entire video into a compact feature representation, learning the semantics and a discriminative feature space;
(b) they are applicable to all kinds of networks like 2D and 3D CNNs for video classification; and (c) they model feature interactions in a more expressive way and without loss of information. We conduct experiments on two challenging human action datasets: HMDB51 and UCF101. The experiments show that TLE outperforms current state-of-the-art methods on both datasets.}
\end{abstract}

\section{Introduction}
\footnote[]{$^{\star}$Ali Diba and Vivek Sharma contributed equally to this work and listed in alphabetical order.}

Human action recognition~\cite{karpathy,twostream,idt} in videos has attracted quite some attention, due to the potential applications in video surveillance, behavior analysis, video retrieval, and more. Even if considerable progress was made, the performance of computer vision systems still falls behind that of people. On top of the challenges that make object class recognition hard, there are issues like camera motion and the continously changing viewpoints that come with it. Whereas Convolutional Networks (ConvNets) have caused several sub-fields of vision to leap forward, they still lack the capacity to exploit long-range temporal information, probably the main reason why end-to-end networks are still unable to outperform methods using hand-crafted features~\cite{idt}. 

\begin{figure}[t]
\centering
{\includegraphics[width=1\columnwidth]{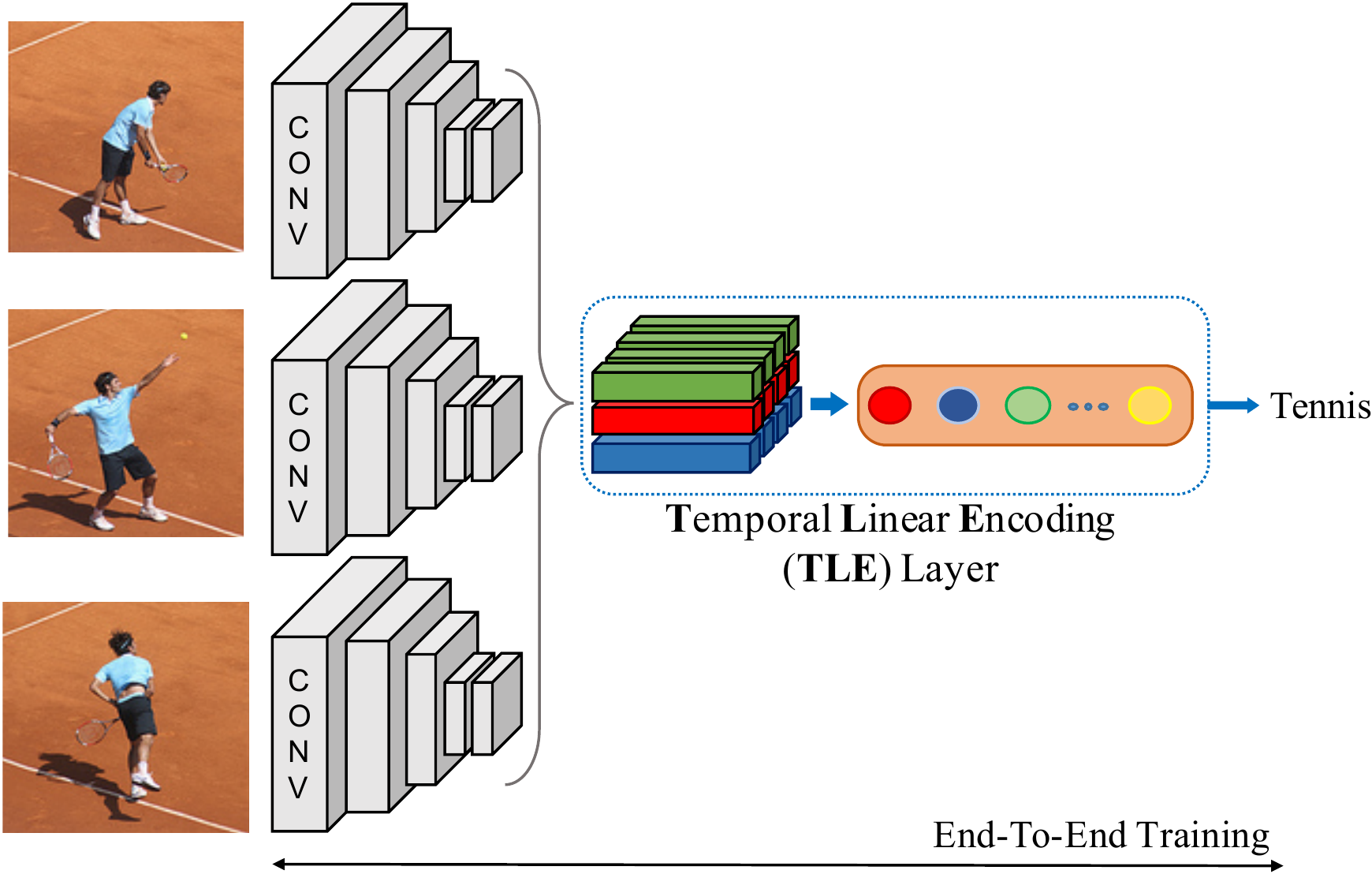} 
} 
\caption{Temporal linear encoding for video classification. Given several segments of an entire video, be it either a number of frames or a number of clips, the model builds a compact video representation from  the spatial and temporal cues they contain, through end-to-end learning. The ConvNets applied to different segments share the same weights.} \label{fig:front}
\end{figure}

Neural networks for action recognition can be categorized into two types, namely \textit{one-stream} ConvNets~\cite{karpathy,c3d} (which use only one stream at a time: either spatial or temporal information), and \textit{two-stream} ConvNets~\cite{twostream} (which integrate both spatial and temporal information at the same time). 

As to the one-stream ConvNets, spatial networks perform action recognition from individual video frames. They lack any form of motion modeling. On the other hand, temporal networks typically get their motion information from dense optical flow. This reliance on dense temporal sampling leads to excessive computational costs for longer videos. One way to avoid processing the abundance of input frames is by extracting a fixed number of shorter clips, evenly distributed over the video~\cite{twostream,c3d}. 

The two-stream ConvNets have shown to outperform one-stream ConvNets. They exploit fusion techniques like trajectory-constrained pooling~\cite{tdd}, 3D pooling~\cite{christopher}, and consensus pooling~\cite{tsn}. The fusion methods of spatial and motion information lie at the heart of the state-of-the-art two-stream ConvNets. 

Motivated by the above observations, we propose the new spatio-temporal encoding illustrated in Figure~\ref{fig:front}. The design of the spatio-temporal deep feature encoding aims to aggregate multiple video segments (i.e. frames or clips) over longer time ranges. To that end, we use our `temporal linear encoding'~(TLE), which is inspired by previous works on video representations~\cite{idt} and feature encoding methods~\cite{bilinearmodels,fishernet}. TLE is a form of temporal aggregation of features sparsely sampled over the whole video using feature map aggregation techniques, and then projected to a lower dimensional feature space using encoding methods powered by end-to-end learning of deep networks. Specifically, TLE captures the important concepts from the long-range temporal structure in different frames or clips, and aggregates it into a compact and robust feature representation by linear encoding. The compact temporal feature representation fits action recognition well, as it is a global feature representation over the whole video. The goal of the paper is not only to achieve high performance, but also to show that TLEs are computationally efficient, robust, and compact.  TLE is evaluated on two challenging action recognition datasets, namely HMDB51~\cite{hmdbdataset} and UCF101~\cite{ucfdataset}. We experimentally show that the two-stream ConvNets when combined with TLEs achieve state-of-the-art performance on HMDB51 (71.1\%) and UCF101 (95.6\%).

The rest of the paper is organized as follows. In Section~\ref{sec:relatedwork}, we discuss related work. Section~\ref{sec:approach} describes our proposed approach. Experimental results and their analysis are presented in Section~\ref{sec:evaluation} and Section~\ref{sec:scene}. Finally, conclusions are drawn in Section~\ref{sec:conclusion}.

\section{Related Work} \label{sec:relatedwork}
\noindent
\textbf{Action Recognition without ConvNets:} Over the last two decades, several action recognition techniques in videos have been proposed by the vision community. Quite a few are concerned with effective representations using local spatio-temporal features, suc h as HOG3D~\cite{hog3d}, SIFT3D~\cite{sift3d}, HOF~\cite{hof}, ESURF~\cite{esurf}, and MBH~\cite{mbh}.  Recently, IDT~\cite{idt} was proposed, which is currently the state-of-the-art among hand-crafted features. Despite this good performance, these features have several shortcomings: they are computationally expensive; they fail to capture semantic concepts; they lack discriminative capacity as well as scalability. To overcome such issues, several techniques have been proposed to model the temporal structure for action recognition, such as the actom sequence model~\cite{actoms} which considers sequence of histograms; temporal action decomposition~\cite{temporalactiondecomposition} which exploits the temporal structure of human actions by temporally decomposing video frames; dynamic poselets~\cite{dynamicposelets} which uses a relational model for action detection; and the temporal evolution of appearance representations~\cite{basura} which uses a ranking function capable of modeling the evolution of both appearance and motion over time.  

\noindent
\textbf{ConvNets for Action Recognition:}
Recently several attempts have been made to go beyond individual image-level appearance information and  exploit the temporal information using ConvNet architectures. End-to-end ConvNets have been introduced in~\cite{christopher,twostream,c3d,tsn} for action recognition. Karpathy et al.~\cite{karpathy} trained a deep network operating on individual frames using a very large sports activities dataset (Sports-1M). Yet, the deep model turned out to be less accurate than an IDTs-based representation because it could not capture the motion information. To overcome this problem, Simonyan et al.~\cite{twostream} proposed a two-stream network, cohorts of spatial and temporal ConvNets. The input to the spatial and temporal networks are RGB frames and stacks of multiple-frame dense optical flow fields, respectively. The network was still limited in its capacity to capture temporal information, because it operated on a fixed number of regularly spaced, single frames from the entire video. Tran et al.~\cite{c3d} explored 3D ConvNets on video streams for spatio-temporal feature learning for clips of 16 frames, and filter kernel of size $3\times 3 \times 3$. In this way, they avoid to calculate the optical flow explicitly and still achieve good performance.  Sun et al.~\cite{sun} proposed a factorized spatio-temporal ConvNet and decomposed the 3D convolutions into 2D spatial and 1D temporal convolutions. Similar to~\cite{twostream} and~\cite{c3d} is Feichtenhofer et al.'s~\cite{christopher} work, where they employ 3D Conv fusion and 3D pooling to fuse spatial and temporal networks using RGB images and a stack of 10 optical flow frames as input. Wang et al.~\cite{tsn} use multiple clips sparsely sampled from the whole video as input for both streams, and then combine the scores for all clips in a late fusion approach.

\noindent
\textbf{Encoding Methods:}
As to prior encoding methods, there is a vast literature on BoW~\cite{bow1,bow2}, Fisher vector encoding~\cite{fv} and sparse encoding~\cite{sparse}. Such methods have performed very well in various vision tasks. FV encoding~\cite{fishernet} and VLAD~\cite{vladnet1,vladnet2} have lately been integrated as a layer in ConvNet architectures, and CNN encoded features have produced superior results for several challenging tasks. Likewise, Bilinear models~\cite{bilinearmodels,oldbilinearmodels} have been widely used and have achieved state-of-the-art results. Bilinear models are computationally expensive as they return matrix outer products, hence can lead to prohibitively high dimensions. To tackle this problem, compact bilinear pooling~\cite{compactbilinear} was proposed which uses the Tensor Sketch Algorithm~\cite{tsalgo}, to project features from a high dimensional space to a lower dimensional one, while retaining state-of-the-art performances. Compact bilinear pooling has shown to perform better than FV encoding and fully-connected networks~\cite{compactbilinear}. Moreover, this type of feature representation is compact, non-redundant, avoids over-fitting, and reduces the number of parameters of CNNs significantly, as it replaces fully-connected layers. 

\begin{figure}[t]
\centering
{\includegraphics[width=0.99\columnwidth]{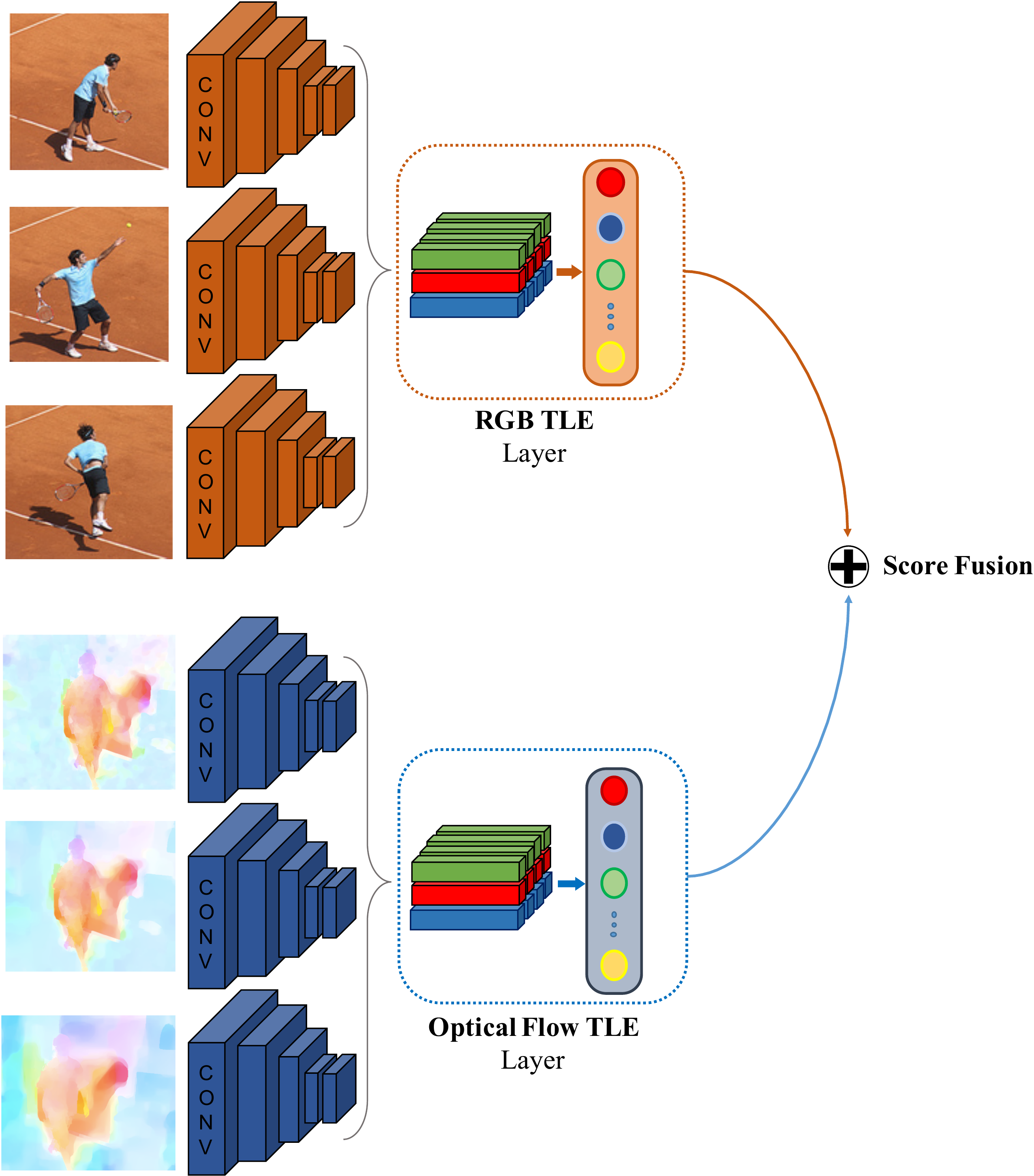} 
} 
\caption{Our temporal linear encoding applied to the design of two-stream ConvNets~\cite{twostream}: spatial and temporal networks. The spatial network operates on RGB frames, and the temporal network operates on optical flow fields. The features maps from the spatial and temporal ConvNets for multiple such segments are aggregated and encoded. Finally, the scores for the two ConvNets are combined in a late fusion approach as averaging. The ConvNet weights for the spatial stream are shared and similarly for the temporal stream.}  \label{fig:main}
\end{figure}

Our proposed temporal linear encoding captures more expressive interactions between the segments across entire videos, and encodes these interactions into a compact representation for video-level prediction. To the best of our knowledge, this is the first end-to-end deep network that encodes  temporal features from entire videos.


\section{Approach} \label{sec:approach}

In a video, the motion between consecutive frames tends to be small. Motivated by this, IDTs~\cite{idt} showed that the densely sampling feature points in video frames and using optical flow to track them yields a good video representation. This suggests that we need a video representation that encodes all the frames together, in order to also capture long-range dynamics. To tackle this issue, recently some techniques have combined several consecutive~\cite{twostream} or sparsely sampled~\cite{tsn} frames into short clips. Unlike IDTs, these techniques use ConvNets with late fusion to combine spatial and temporal cues, but they still fail to efficiently encode all frames together.

\begin{algorithm}[b]
\caption{Deep Temporal Linear Encoding Layer}\label{algo:the_alg}
\begin{algorithmic}
\STATE \textbf{Input:} CNN features for $K$ frames/clips \{$S_{1},S_{2},...,S_{K}$\} of video $V$, $S \in \mathbb{R}^{ h \times w \times c}$, where $h$, $w$ and $c$ are height, width, and channels of feature maps respectively.
\STATE \textbf{Output:} Temporal linear encoded feature map $y \in \mathbb{R}^{d}$, where $d$ is the encoded feature dimension.
\STATE \textbf{Temporal Linear Encoding:}
\STATE \textbf{1.} $X=S_{1} \diamond S_{2} \diamond \ldots \diamond S_{K}$, $X \in \mathbb{R}^{ h \times w \times c}$, where $\diamond$ is an aggregation operator
\STATE \textbf{2.} $y=EncodingMethod(X)$, $y \in \mathbb{R}^{d}$, where $d$ denotes the encoded feature dimensions
\end{algorithmic}
\end{algorithm}

\begin{figure}[t]
\centering
{\includegraphics[width=0.99\columnwidth]{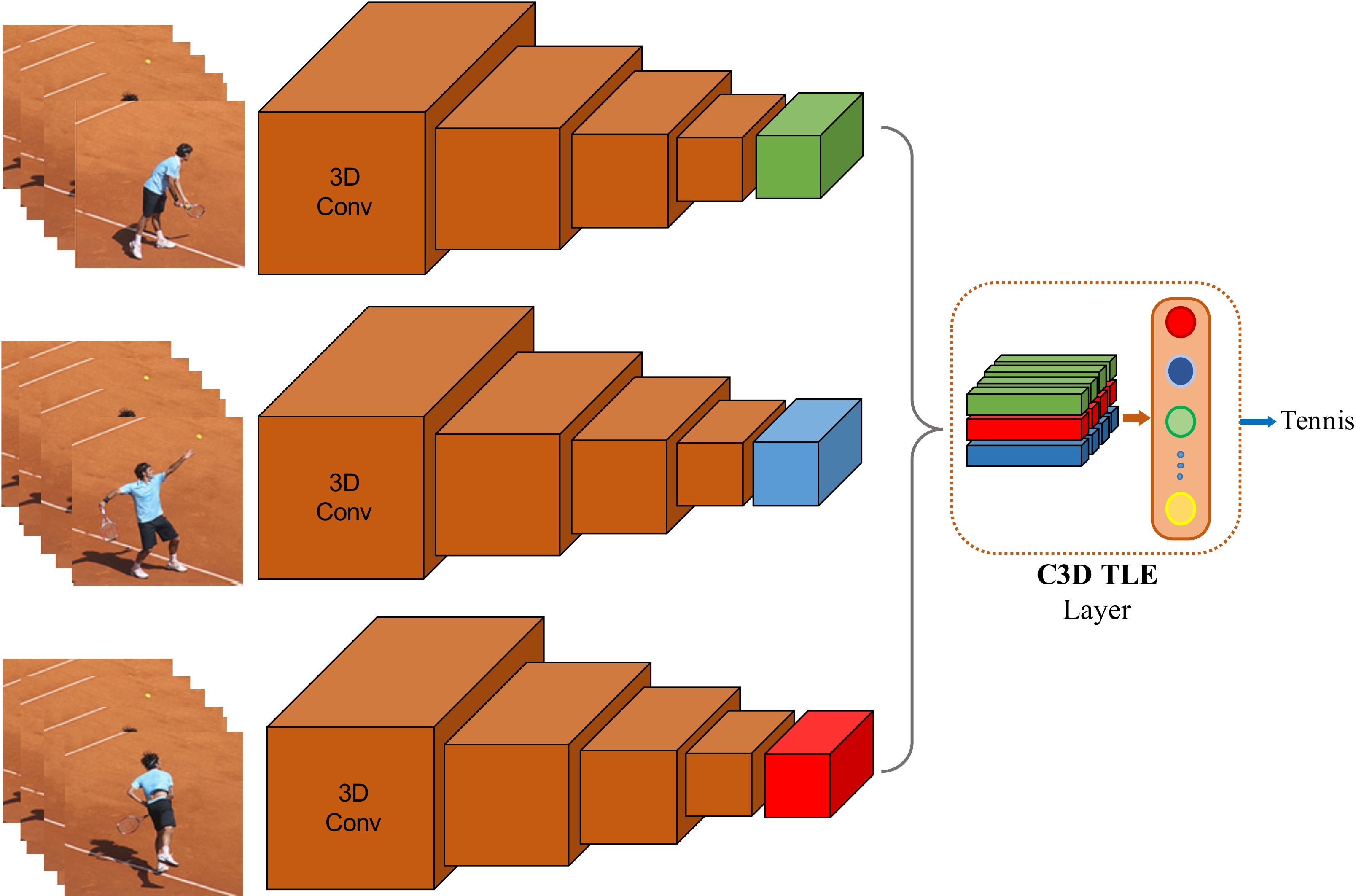} 
} 
\caption{Our temporal linear encoding applied to 3D ConvNets~\cite{c3d}. These use video clips as input. The feature maps from the clips are aggregated and encoded. The output of the network is a video-level prediction. The ConvNets operating on the different clips all share the same weights.}  
\label{fig:main2}
\end{figure}

Given earlier successes with deep learning, creating effective video representations should seem possible via the end-to-end learning of deep neural networks. The hope would be that such representations embody more of the semantic information extracted along the whole video. Our goal is to create a single feature space in which to represent each video using all its selected frames or clips, rather than scoring separate frames/clips with classifiers and label the video based on scores aggregation. We propose temporal linear encoding (TLE) to aggregate spatial and temporal information from an entire video, and to encode it into a robust and compact representation, using end-to-end learning, as shown in Fig.~\ref{fig:main} and Fig.~\ref{fig:main2}.  Algorithm~\ref{algo:the_alg} sketches the steps of the proposed TLE. More details about the CNN encoding layer is given in Section~\ref{subsec:tle}.

\subsection{Deep Temporal linear encoding} \label{subsec:tle}
Consider the output feature maps of CNNs truncated at a convolutional layer for $K$ segments extracted from a video $V$. The feature maps are matrices \{$S_{1},S_{2},...,S_{K}$\} of  size $S \in \mathbb{R}^{h \times w \times c}$, where $h$, $w$ and $c$ denote the height, width, and number of channels of the CNN feature maps. A temporal aggregation function $T : S_{1},S_{2},\ldots,S_{K} \rightarrow X$, aggregates $K$ temporal feature maps to output an encoded feature map $X$. The aggregation function can be applied to the output of different convolutional layers. This temporal aggregation allows us to linearly encode and aggregate information from the entire video into a compact and robust feature representation. This retains the temporal relationship between all the segments without the loss of important information. We investigated different functions $T$ for the temporal aggregation of the segments.
\begin{itemize}
\item Element-wise Average of Segments:
\begin{equation}
X=(S_{1} \oplus S_{2} \oplus \ldots \oplus S_{K})/K
\end{equation}
\item Element-wise Maximum of Segments:
\begin{equation}
X=max\{S_{1},S_{2},\ldots,S_{K}\} 
\end{equation}
\item Element-wise Multiplication of Segments:
\begin{equation}
X=S_{1} \circ S_{2} \circ \ldots \circ S_{K}
\end{equation}
\end{itemize}
Of all the temporal aggregation functions illustrated above, element-wise multiplication of feature maps yielded best results, and was therefore selected.

The temporal aggregated matrix $X$ is fed as input to an encoding (or pooling) method $E: X\rightarrow y$, resulting in a linearly encoded feature vector $y, y \in \mathbb{R}^{d}$, where $d$  denotes the encoded feature dimensions. The advantage of encoding is that every channel of the aggregated temporal segments interacts with every other channel, thus leading to a powerful feature representation of the entire video. In this work, we investigate two encoding methods $E$:

\begin{itemize}
\item \textbf{Bilinear Models}:  A bilinear model~\cite{bilinearmodels,oldbilinearmodels} computes the outer product of two feature maps, given by:
\begin{equation}
y=W[X \otimes X^{'}]
\end{equation}
Where $X \in \mathbb{R}^{(hw)\times c}$, $X^{'} \in \mathbb{R}^{(hw)\times c^{'}}$ are input feature maps, $y \in \mathbb{R}^{(c c^{'})}$ are bilinear features, $\otimes$ denotes the outer product, [ ] turns the matrix into a vector by concatenating the columns, and $W$ represents model parameters to be learned (here linear). In our case, $X=X^{'}$. The resulting bilinear features capture the interaction of features with each other at all spatial locations, hence leading to a high-dimensional representation. For this reason, we use the Tensor Sketch algorithm~\cite{compactbilinear,tsalgo},  which projects this high-dimensional space to a lower-dimensional space, without computing the outer product directly. That cuts down on the number of model parameters significantly. The model parameters $W$ are learned with end-to-end back propagation. 

\item \textbf{Fully connected pooling}: As the network has fully-connected layers between the last convolutional layer and the classification layer, the model parameters of the fully-connected layer and classification layer are learned when training the network from scratch or when fine-tuning a pre-trained network.
\end{itemize}

Compared to the fully-connected pooling method, bilinear models projects the high dimensional feature space to a lower dimensional space, which is far fewer in parameters  and still perform better than fully-connected layers in performance, apart from computational efficiency.

One can readily employ other encoding methods like deep fisher encoding~\cite{fishernet} or VLAD~\cite{vladnet1,vladnet2}, instead of bilinear models or fully connected pooling. When bilinear models are used the features are passed through a signed squared root and $L_{2}$-normalization. In either case, we use softmax as a classifier.

\begin{figure}[t]
\centering
{\includegraphics[width=0.99\columnwidth]{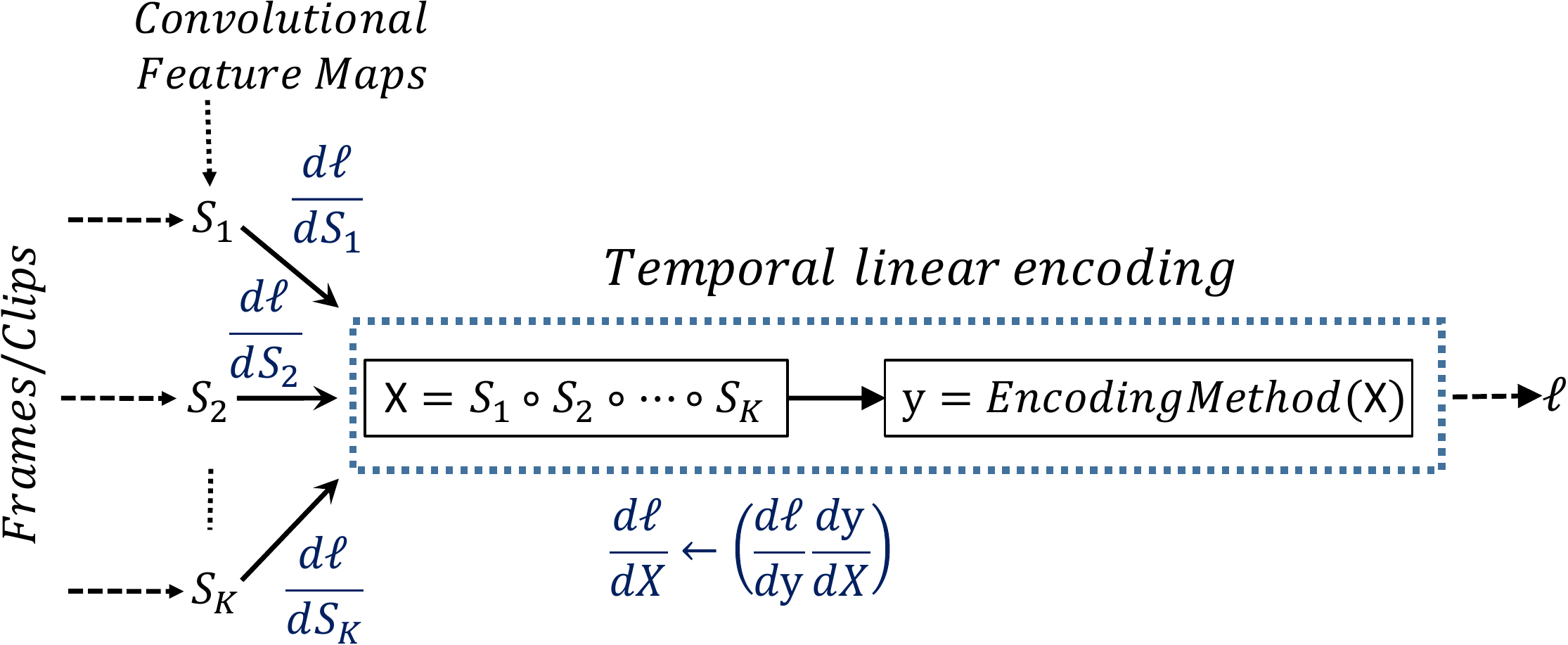} 
} 
\caption{Computing gradients for back-propagation in the temporal linear encoding.} \label{fig:gradient}
\end{figure}

\textbf{End-to-end training:} 
We use $K=3$, following the advice from temporal modeling work~\cite{actoms}. Let the output feature maps of the CNNs be $S_{1}$, $S_{2}$, and $S_{3}$. The temporally aggregated features are given by $X=S_{1} \circ S_{2} \circ S_{3}$, and the temporal linearly encoded features are denoted by $y$. Let $\ell$ denote the loss function, and $d\ell/d(X)$ be the gradient of the loss function with respect to $X$. Algorithm~\ref{algo:the_alg_examples} illustrates the forward and backward passes of our temporal linear encoding steps for the 3 segments setup.

The Back-propagation for the joint optimization of the $K$ temporal segments can be derived as:
\begin{equation}
\begin{split}
\frac{d\ell}{dS_{k}}=((S_{1} \circ...\circ S_{K}) \backslash S_{k}) \frac{d\ell}{dX}, k \in K  
\end{split}
\label{eq:5}
\end{equation}

In the end-to-end learning, the model parameters for the $K$ temporal segments are optimized using stochastic gradient descent (SGD). Moreover, the temporal linear encoding model parameters are learned from the entire video. The scheme is illustrated in Fig.~\ref{fig:gradient}.

\begin{algorithm}[t]
\caption{Forward \& backward propagation steps for our deep temporal linear encoding with bilinear models for a scheme with 3 segments.}\label{algo:the_alg_examples}
\begin{algorithmic}
\STATE \textbf{Input:} Convolutional feature maps for a scheme of 3 segments, \{$S_{1},S_{2},S_{3}$\}, $S \in \mathbb{R}^{ h \times w \times c}$
\STATE \textbf{Output:} $y \in \mathbb{R}^{d}$
\STATE \textbf{Temporal Linear Encoding:}
\STATE \textbf{Forward Pass:} 
\STATE \textbf{1.}  $X=S_{1} \circ S_{2} \circ  S_{3}$, $X \in \mathbb{R}^{ h \times w \times c}$
\STATE \textbf{2.} $y=[XX^{T}]$, $y \in \mathbb{R}^{c^{2}}$
\STATE \textbf{Backward Pass:} 
\STATE \textbf{1.}   
$\frac{d\ell}{dS_{1}}=(S_{2} \circ S_{3}) \frac{d\ell}{dX}$,  \newline $\frac{d\ell}{dS_{2}}=(S_{1} \circ S_{3}) \frac{d\ell}{dX}$, \newline $\frac{d\ell}{dS_{3}}=(S_{1} \circ S_{2}) \frac{d\ell}{dX}$
\end{algorithmic}
\end{algorithm}

\section{Evaluation} \label{sec:evaluation}

In this section, we first introduce the datasets and implementation details of our proposed approach. Then we demonstrate the applicability of our temporal linear encoding on 2D and 3D ConvNets using frames or clips  to encode long-range dynamics across entire videos. Finally, we compare temporal linear encoding with the state-of-the-art methods.

\subsection{Datasets}
We conduct experiments on two challenging video datasets with human actions, namely HMDB51~\cite{hmdbdataset} and UCF101~\cite{ucfdataset}. The HMDB51 dataset consists of 51 action categories with 6,766 video clips in all. The UCF101 dataset consists of 101 action classes with 13,320 video clips. Both of these datasets have at least 100 video clips for each action category. For both datasets, we use the three training/testing splits provided as the original evaluation scheme for these datasets, and report the average accuracy over these three splits.  

\subsection{Implementation details}

We use the caffe toolbox~\cite{caffe} for ConvNet implementation and all the networks are trained on two Geforce Titan X GPUs. Here, we describe the implementation details of our two schemes, temporal linear encoding with two-stream ConvNets and temporal linear encoding with C3D ConvNets using bilinear models and fully-connected pooling. As mentioned earlier in the approach section, we use 3 segments for ConvNet training and testing.

\paragraph*{Two-stream ConvNets:} We employ three pre-trained models trained on the ImageNet dataset~\cite{imagenet}, namely AlexNet~\cite{alexnet}, VGG-16~\cite{vgg}, and BN-Inception~\cite{inception}, for the design of the two-stream ConvNets. The two-stream network consists of spatial and temporal networks, the spatial ConvNet operates on RGB frames, and the temporal ConvNet operates on a stack of $10$ dense optical flow frames. The input RGB image or optical flow frames are of size $256 \times 340$, and  are randomly cropped to a size $224 \times 224$,  and then mean-subtracted for network training. To fine-tune the network, we replace the previous classification layer with a $C$-way softmax layer, where $C$ is the number of action categories.  We use mini-batch stochastic gradient descent (SGD) to learn the model parameters with a fixed weight decay of $5 \times 10^{-4}$, momentum of 0.9, and a batch size of 15 for network training.  The prediction scores of the spatial and temporal ConvNets are combined in a late fusion approach as averaging before softmax normalization.

\textbf{$-$ TLE with Bilinear Models:} In our experiments for bilinear models, we retain only the convolutional layers of each network, more specifically we remove all the fully connected layers, similar to~\cite{compactbilinear,bilinearmodels}. The convolutional feature maps extracted from the last convolutional layers (after the rectified output of the last convolutional layer, when there is one) are fed as input into the bilinear models. For example, the convolutional feature maps for the last layer of BN-Inception produces an output of size $14 \times 14 \times 1024$, leading to bilinear features $1024 \times 1024$, and 8,196 features for compact bilinear models.  We follow two steps to fine-tune the whole model. First, we train the last layer using logistic regression. Secondly, we fine-tune the whole model. In both steps for training spatial ConvNets, we initialize the learning rate with $10^{-3}$ and decrease it by a factor of 10 every 4,000 iterations. The maximum number of iterations is set to 12,000. We use flip augmentation about the horizontal axis and RGB jittering for RGB frames. For the temporal ConvNet, we use a stack of $10$ optical flow frames as input clip. We rescale the optical flow fields linearly to a range of [$0,255$] and compress as JPEG images. For the extraction of the optical flow frames, we use the TVL1 optical flow algorithm~\cite{opticalflow} from the OpenCV toolbox with CUDA implementation. In both steps for training the temporal ConvNets, we initialize the learning rate with $10^{-3}$ and manually decrease by a factor of 10 every 10,000 iterations. The maximum number of iterations is set to 30,000. We use batch normalization. Before the  features are fed into the softmax layer, the features are passed through a signed squared root operation ($z \leftarrow sign(y)\sqrt{|y|}$)  and $L_{2}$-normalization ($z^{'} \leftarrow z/||z||_{2}$). 

\textbf{$-$ TLE with Fully-Connected Pooling:} We follow the same two step fine-tuning scheme discussed earlier. For fine-tuning the fully-connected layers between the last convolutional layer and the $C$-way softmax layer for both spatial and temporal ConvNets, we initialize the learning rate with $10^{-3}$ and decrease it by a factor of 10 every 10,000 iterations in both model training steps. The maximum number of iterations is set to 30,000. We apply the same augmentation and discretization techniques for both RGB and optical flow frames, as discussed earlier.

\paragraph*{C3D ConvNets:}  In our experiments, we use the C3D model~\cite{c3d} pre-trained on the  Sport-1M dataset~\cite{karpathy}. The convolution kernels are of size $3 \times 3 \times 3$ with stride 1 in both spatial and temporal dimensions, as suggested in~\cite{c3d}. The video is decomposed into non-overlapping, equal-duration clips of 16 frames. The C3D ConvNet operates on these video clips as input for network training. The video frames are of size $128 \times 171$. For network training, we randomly crop the video clips to a size $16 \times 112 \times 112$,  and then mean-subtract. We use a single center crop per clip. For fine-tuning the network, we replace the previous classification layer with a $C$-way softmax layer, where $C$ is the number of action categories.  We use mini-batch stochastic gradient descent to learn the model parameters with a fixed weight decay of $5 \times 10^{-4}$, momentum of 0.9, and a batch size of 10 for network training.

Also in this work, following the same fine-tuning scheme as the original C3D ConvNets on UCF101, we fine-tune the C3D ConvNets on HMDB51 and report their average accuracy over the three splits.

\textbf{$-$ TLE with Bilinear Models:} Similar to TLE with bilinear models from two-stream ConvNets, we retain the  convolutional layers. For fine-tuning the model, we use the same two steps scheme explained earlier. In both steps for C3D ConvNets training, we initialize the learning rate with $3\times 10^{-3}$ and decrease by a factor of 10 every 10,000 iterations. The maximum number of iterations is set to 30,000. We use batch normalization. Before feeding the features to the softmax classifier, the features are passed through a signed squared root and  $L_{2}$-normalization.

\textbf{$-$ TLE with Fully-Connected Pooling:} To fine-tune the fully-connected layers of the C3D ConvNets, we follow the same two-step fine-tuning scheme discussed earlier. In both steps of model training, we initialize the learning rate with $10^{-3}$ and manually decrease by a factor of 10 every 10,000 iterations. The maximum number of iterations is set to 40,000.

\paragraph*{Testing:} \emph{}

\textbf{$-$ Two-stream ConvNets:} 
Given a video, we divide it into 3 parts of equal duration. The three parts are associated with the 3 segments. For TLE two-stream ConvNet testing, at a time, we extact 1 RGB frame or 10 optical flow frames from each part and feed these into the 3 segments network sequentially. In total, we sample 5 RGB frames or stacks of optical flow frames (i.e. 15 frames for the three-segments in total) from the whole video.   For video prediction, we average predictions over all group of frame segments.  The prediction scores of the spatial and temporal ConvNets are combined in a late fusion approach via averaging.

\textbf{$-$ C3D ConvNets:} 

We decompose each video into non-overlapping clips of 16 frames, we then divide the number of clips into 3 equal parts. For TLE C3D ConvNets testing, 1 clip is extracted from each part and fed sequentally into the 3 segment network. In total, we extract 3 clips (i.e. 9 clips for three-segments in total) from the whole video. We average the predictions over all groups of clip segments to make a video-level prediction.

\subsection{Evaluation of TLE}

In this subsection, we explore  (i) different aggregation functions $T$ to linearly aggregate the segments into  a  compact intermediate representation for encoding; and (ii) different ConvNet architectures for  both two-stream (spatial and temporal networks) and C3D networks. For this evaluation, we report the accuracy of split1 on UCF101 and HMDB51. The reported performance is for TLE with bilinear models using the tensor sketch algorithm.

\paragraph*{Two-Stream ConvNets:} \emph{}

\textbf{$-$ Aggregation Function:} In our evaluation, we explore three aggregations functions (i) element-wise average, (ii) element-wise maximum, and (iii) element-wise multiplication. In Table~\ref{table:2D_agg}, we report the performance of the different aggregation strategies. We observe that the element-wise multiplication performs the best.
Therefore, we choose element-wise multiplication as a default aggregation function. We believe combining the feature maps in this way allows us to aggregate the appearance and motion information accurately, hence leading to better results. Interestingly, we also found that aggregating rectified output of the last convolutional feature maps achieves around the same classification performance, in comparison to non-rectified ones.

\begin{table}[htb] 
\begin{center}
\resizebox{7cm}{!} {
\begin{tabular}{| l |   c |   }
\hline
Aggregation Function ($T$)  			& 		UCF101/HMDB51\\
\hline
Element-wise Maximum			&  91.3/67.4	\\ \hline
Element-wise Average			&  92.6/68.1	\\ \hline
Element-wise Multiplication		&  \textbf{94.8/70.4} 	\\ \hline 
\end{tabular}}
\end{center} \vspace{-0.2cm}
\caption{Accuracy~(\%) performance comparison of the aggregation functions in {TLE} BN-Inception network for {two-stream ConvNets} using 3 segments on UCF101 and HMDB51 datasets (split1).}
\label{table:2D_agg}
\end{table}

\textbf{$-$ ConvNet Architectures:} Here, we compare the different ConvNet architectures for TLE. Specifically, we compare AlexNet~\cite{alexnet}, VGG-16~\cite{vgg}, and BN-Inception~\cite{inception}. Among all architectures shown in Table~\ref{table:2D_spat_temp}, BN-Inception achieves the best performance, better than the AlexNet and VGG-16 architectures. BN-Inception is 5.4/2.3\% (spatial ConvNets) and 2.3/4.9\% (temporal ConvNets) better than VGG-16 on UCF101/HMDB51. Therefore, we choose BN-Inception as a default ConvNet architecture for TLE. We can observe that the deeper models, the higher is the performance gain on both datasets.

\begin{table}[t] 
\begin{center}
\resizebox{7.5cm}{!} {
\begin{tabular}{| l |   c |  c |  }
\hline
 &  UCF101/HMDB51 & UCF101/HMDB51\\
\hline
Method &  Spatial ConvNets & Temporal ConvNets\\
\hline
AlexNet			&  74.4/50.8&  82.7/52.4	\\ \hline
VGG-16			&  81.5/60.9&  86.8/61.5\\ \hline
BN-Inception	&  \textbf{86.9/63.2}&  \textbf{89.1/66.4}\\ \hline 
\end{tabular}}
\end{center}\vspace{-0.2cm}
\caption{Different architecture accuracy~(\%) performance comparison of spatial and temporal ConvNets using 3 segments on the UCF101 and HMDB51 datasets (split1).}
\label{table:2D_spat_temp}
\end{table}

\paragraph*{C3D ConvNets:} \emph{}

\textbf{$-$ Aggregation Function:}  We perform similar experiments to explore the aggregations functions in C3D ConvNets, as used in two-stream ConvNets. Table~\ref{table:3D_agg} summarizes the results of comparing different aggregation strategies. Similar to two-stream ConvNets, element-wise multiplication performs better in comparison to other candidate functions, and was therefore selected as a default aggregation function.

\begin{table}[htb] 
\begin{center}
\resizebox{7cm}{!} {
\begin{tabular}{| l | c |}
\hline
Aggregation Function ($T$)  			& 		Accuracy (\%)\\
\hline
Element-wise Maximum			&   84.2	\\ \hline
Element-wise Average			&   84.6	\\ \hline
Element-wise Multiplication		&   \textbf{86.1}	\\ \hline 
\end{tabular}}
\end{center}\vspace{-0.2cm}
\caption{Performance comparison of different aggregation functions in {TLE} C3D ConvNet using 3 segments on the UCF101 dataset (split1).}
\label{table:3D_agg}
\end{table}

\textbf{$-$ ConvNet Architectures:}  We use the C3D ConvNet~\cite{c3d} architecture as a default ConvNet architecture for TLE. The model obtains an accuracy of 86.1\% for  split1 on the UCF101 dataset.

\subsection{Comparison with the state-of-the-art}
Finally, after exploring the aggregation function and good ConvNet architectures, we compare our TLE with the current state-of-the-art methods over all three splits of the UCF101 and HMDB51 datasets.  We report the average accuracy over the three splits of both datasets.

\begin{table}[b] 
\begin{center}
\resizebox{7.4cm}{!} {
\begin{tabular}{| l |   c |  c |  }
\hline
Method &  UCF101 & HMDB51\\
\hline
DT+MVSM~\cite{dtmvsv}			& 83.5 & 55.9	\\ \hline
iDT+FV~\cite{idt}			& 85.9 & 57.2	\\ \hline
Two Stream~\cite{twostream}		& 88.0 & 59.4	\\ \hline
VideoDarwin~\cite{basura}		& $-$ &	63.7	\\ \hline
C3D~\cite{c3d}	& 82.3 & 56.8	\\ \hline
Two Stream+LSTM~\cite{twostreamlstm}	& 88.6 & $-$	\\ \hline
F$_{ST}$CV (SCI fusion)~\cite{sun}	& 88.1 & 59.1	\\ \hline
TDD+FV~\cite{tdd}		& 90.3 &	63.2	\\ \hline
LTC~\cite{varol2016long}			& 91.7 &	64.8	\\ \hline
KVMF~\cite{KVMF}		& 93.1 &	63.3	\\ \hline
TSN~\cite{tsn}	& 94.0 & 68.5	\\ \hline
3DConv+3DPool~\cite{christopher}			& 93.5 & 69.2	\\ \hline
\hline
{TLE: FC-Pooling (ours)}			& {92.2}		&  {68.8}	\\ 
\hline
{TLE: Bilinear+TS (ours)}			& {95.1}		&  {70.6}	\\ 
\hline
\textbf{TLE: Bilinear (ours)}			& \textbf{95.6}		&  \textbf{71.1}	\\ 
\hline
\end{tabular}}
\end{center} \vspace{-0.2cm}
\caption{\textbf{Two-stream ConvNets.} Accuracy~(\%) performance comparison of TLE BN-Inception network with state-of-the-art methods over all three splits of UCF101 and HMDB51.}
\label{table:soa1}
\end{table}

\textbf{$-$ Two-stream ConvNets:} In Table~\ref{table:soa1}, we compare the performance of TLE with the current methods using two-stream ConvNets and other traditional methods. TLE with bilinear models (TLE:Bilinear) performs the best among all methods. This model obtains an accuracy of 95.6\% and 71.1\% on UCF101 and HMDB51, respectively. The performance gap of TLE with bilinear models using the  tensor sketch algorithm (TLE:Bilinear+TS) is however small 0.5/0.5\%, and  3.4/2.3\% for TLE with fully-connected pooling (TLE:FC-Pooling) when compared to TLE:Bilinear on UCF101/HMDB51. TLE:Bilinear is 7.6/11.7\%, 1.6/2.6\%, and 2.1/1.9\% better than the Two-Stream~\cite{twostream}, TSN~\cite{tsn}, and 3DConv+3DPool~\cite{christopher} methods on UCF101/HMDB51.  One can observe that the optical flow is better at capturing the motion information (shown in Table~\ref{table:2D_spat_temp}), and when combined with the appearance information in long-range temporal structure is effective to perform video-level learning. As another interesting comparison, our TLE with bilinear models yields very few parameters to train, in comparison to other methods which have several fully-connected layers to train with millions of parameters. Thus, our models are computationally efficient. Moreover, our models clearly show the power of encoded feature representations in video classification for entire videos, in end-to-end learning.

\textbf{$-$ C3D ConvNets:} In Table~\ref{table:soa2}, we summarize the performance of TLE in C3D ConvNets and compare it with the currently used 3D Conv~\cite{c3d},  and other traditional methods. Similar to two-stream ConvNets, TLE:Bilinear  outperforms other methods, and achieves an accuracy of 86.3\% and 60.3\% on UCF101 and HMDB51, respectively, which is 4/3.5\%, and 0.4/3.1\% better than the original C3D ConvNets~\cite{c3d} and iDT+FV~\cite{idt} methods on UCF101/HMDB51. The goal of this experiment is to show that TLEs can improve the performance of the original C3D ConvNets~\cite{c3d}. It is also interesting to see that, TLE:Bilinear improves the C3D ConvNet performance over the  two-stream ConvNets~\cite{twostream} on the HMDB51 dataset. We think the reason for TLE:Bilinear performing better than other methods is that the model is essentially able to encode the dynamic appearance and motion using multiple aspects of long-range temporal cues in the video data, which were unavailable to the original C3D ConvNets~\cite{c3d}. 

\begin{table}[t] 
\begin{center}
\resizebox{7.4cm}{!} {
\begin{tabular}{| l |  c |  c |  }
\hline
Method &  UCF101 & HMDB51\\
\hline
SpatioTemporal ConvNet~\cite{karpathy}	& 65.4 & $-$	\\ \hline
LRCN~\cite{lrcn}					& 82.9 & $-$	\\ \hline
Composite LSTM Model~\cite{compositelstm}	& 84.3 & 44.0	\\ \hline 
iDT+FV~\cite{idt}			& 85.9 & 57.2	\\ \hline
Two Stream~\cite{twostream}		& \textbf{88.0} & {59.4}	\\ \hline
C3D~\cite{c3d}						& 82.3 & 56.8	\\ \hline
\hline
{TLE: FC-Pooling (ours)}			& {83.1}		&  {58.6}	\\ 
\hline
{TLE: Bilinear+TS (ours)}				&	{85.6}	&  {59.7}\\
\hline
\textbf{TLE: Bilinear (ours)}			& \textbf{86.3}		&  \textbf{60.3}	\\ 
\hline

\end{tabular}}
\end{center} \vspace{-0.2cm}
\caption{\textbf{C3D ConvNets.} Accuracy~(\%) performance comparison of TLE with state-of-the-art methods over all three splits of UCF101 and HMDB51.}
\label{table:soa2}
\end{table}


\section{Scene Context Embedding} \label{sec:scene}

This section describes an additional experiment to incorporate scene context in order to improve the success of action recognition.

ConvNets trained on individual frames in spatial networks tend to misclassify the contextual information from scenery and objects in videos, which could be an evident source of information for action recognition. As an application of our method, we study to  incorporate the context information from the scenes to improve the action recognition performance. 
Our network architecture uses the TLE with 3 segments. In addition we add a fourth segment ConvNet pre-trained on the Places365 dataset~\cite{places365}. The key reason behind using the latter is to supervise the learning of extra representations of scene-related information, to further boost the action recognition. In this way, we transfer the learned representations between the two tasks for better action recognition. We are aware that in this case we use additional data, but it is a nice way to demonstrate the capability of TLE to combine different data streams. 

In this experiment, we  exploit the context information from the scenes to improve the action recognition. 
We apply the same training scheme for two-stream ConvNets explained in Section 4.2.  We use the VGG-16 network architecture in this experiment. In Table~\ref{table:2D_multi}, we observe that the action recognition accuracy of this proposed method outperforms the TLE method with three segments (as shown in Table~\ref{table:2D_spat_temp}) on spatial ConvNets. The result indicates that the two streams of information are encoded as complementary information.

\begin{table}[t] 
\begin{center}
\resizebox{8cm}{!} {
\begin{tabular}{| l |   c |  }
\hline
 &  UCF101/HMDB51 \\
\hline
Method &  Spatial ConvNets \\
\hline
TLE:Bilinear+TS,Action (ours)			& 81.5 / 60.9\\ \hline
\textbf{TLE:Bilinear+TS,Action+Context (ours)}			&  \textbf{83.8 / 63.6}\\ \hline
\end{tabular}}
\end{center}\vspace{-0.2cm}
\caption{Accuracy~(\%) performance comparison of the VGG-16 spatial ConvNets using 3 segments, when combined with context information pre-trained on Places365~\cite{places365}. The accuracy is reported for split1 on both datasets. }
\label{table:2D_multi}
\end{table}

\section{Conclusion} \label{sec:conclusion}

In this paper, we proposed Temporal Linear Encoding  (TLE) embedded inside ConvNet architectures, aiming to aggregate information from an entire video, be it in form of frames or clips. The result is a global feature representation obtained in an end-to-end learning scheme. The model performs action prediction over an entire video. We demonstrated the TLE on two challenging action video datasets: HMDB51 and UCF101. In addition to yielding a better performance than the state-of-the-art methods, TLEs are computationally efficient, robust, compact, reduce the number of model parameters significantly below that of fully-connected ConvNets, and retain the feature interaction in a more expressive way without an undesired loss of information. Even though, in this paper, we have focused on two-stream and C3D ConvNets architectures, our method has the potential to generalize to other architectures, and can readily be employed with other encoding methods also. Thus, it can lead to more accurate classification. Another potential of this work is that TLEs are flexible enough to be readily employed to other forms of sequential data streams for feature embedding.

In future work, concerning  the spatial and temporal segment aggregation, we plan to further investigate architectural alternatives. For instance, one could combine the spatial and temporal networks for each segment individually and then aggregate this spatio-temporal network in hierarchical fashion for a global spatio-temporal feature representation.

\subsection*{Acknowledgements}
This work was supported by DBOF PhD scholarship, KU Leuven CAMETRON project. The authors would like to thank Nvidia for GPU donation.

{
\small
\bibliographystyle{ieee}
\bibliography{egbib}
}

\end{document}